\documentclass[journal]{IEEEtran}
\usepackage{amsmath,amsfonts}
\usepackage{algorithmic}
\usepackage{algorithm}
\usepackage{array}
\usepackage[caption=false,font=normalsize,labelfont=sf,textfont=sf]{subfig}
\usepackage{textcomp}
\usepackage{stfloats}
\usepackage{url}
\usepackage{verbatim}
\usepackage{graphicx}
\usepackage{enumitem}
\usepackage{xcolor}
\usepackage{cite}
\usepackage{pifont} 
\usepackage{graphicx} 
\usepackage{xcolor,colortbl} 
\usepackage{tabularx}
\usepackage{booktabs}
\usepackage{multirow}
\usepackage{hyperref}
\hypersetup{hypertex=true,
colorlinks=true,
linkcolor=blue,
anchorcolor=blue,
citecolor=blue}
\begin{document}

\title{Video-to-Task Learning via Motion-Guided Attention for Few-Shot Action Recognition}


\author{Hanyu Guo, Wanchuan Yu, Suzhou Que, Kaiwen Du,\\ Yan Yan,~\IEEEmembership{Senior Member,~IEEE}, and Hanzi Wang,~\IEEEmembership{Senior Member,~IEEE}%
    \thanks{This work was supported in part by the National Key Research and Development Program of China under Grant 2022ZD0160402; and in part by the National Natural Science Foundation of China under Grant U21A20514, Grant 62372388 and Grant 62071404. (Corresponding author: Hanzi Wang.)}%
    \thanks{Hanyu Guo, Wanchuan Yu, Suzhou Que, Kaiwen Du and Yan Yan are with the Fujian Key Laboratory of Sensing and Computing for Smart City, School of Informatics, Xiamen University, Xiamen 361005, China (email: guohanyu@stu.xmu.edu.cn; wanchuan@stu.xmu.edu.cn; suzhouque@stu.xmu.edu.cn;  kaiwendu@stu.xmu.edu.cn; yanyan@xmu.edu.cn).}%
    \thanks{Hanzi Wang is with the Fujian Key Laboratory of Sensing and Computing for Smart City, School of Informatics, Xiamen University, Xiamen 361005, China, and also with the Shanghai Artificial Intelligence Laboratory, Shanghai 200232, China (e-mail: hanzi.wang@xmu.edu.cn).}}

\markboth{\textit{Guo et al.}: Video-to-Task Learning via Motion-Guided Attention for Few-Shot Action Recognition}%
{Shell \MakeLowercase{\textit{et al.}}: A Sample Article Using IEEEtran.cls for IEEE Journals}

\maketitle

\begin{abstract}
In recent years, few-shot action recognition has achieved remarkable performance through spatio-temporal relation modeling. Although a wide range of spatial and temporal alignment modules have been proposed, they primarily address spatial or temporal misalignments at the video level, while the spatio-temporal relationships across different videos at the task level remain underexplored. Recent studies utilize class prototypes to learn task-specific features but overlook the spatio-temporal relationships across different videos at the task level, especially in the spatial dimension, where these relationships provide rich information. In this paper, we propose a novel Dual Motion-Guided Attention Learning method (called DMGAL) for few-shot action recognition, aiming to learn the spatio-temporal relationships from the video-specific to the
task-specific level. To achieve this, we propose a carefully designed Motion-Guided Attention (MGA) method to identify and correlate motion-related region features from the video level to the task level. Specifically, the Self Motion-Guided Attention module (S-MGA) achieves spatio-temporal relation modeling at the video level by identifying and correlating motion-related region features between different frames within a video. The Cross Motion-Guided Attention module (C-MGA) identifies and correlates motion-related region features between frames of different videos within a specific task to achieve spatio-temporal relationships at the task level. This approach enables the model to construct class prototypes that fully incorporate spatio-temporal relationships from the video-specific level to the task-specific level. We validate the effectiveness of our DMGAL method by employing both fully fine-tuning and adapter-tuning paradigms. The models developed using these paradigms are termed DMGAL-FT and DMGAL-Adapter, respectively. Extensive experiments demonstrate that both our DMGAL-FT and DMGAL-Adapter models achieve favorable results compared to other state-of-the-art methods on five few-shot action recognition benchmarks.
\end{abstract}

\begin{figure}[t]
	\centering
	\includegraphics[width=\linewidth]{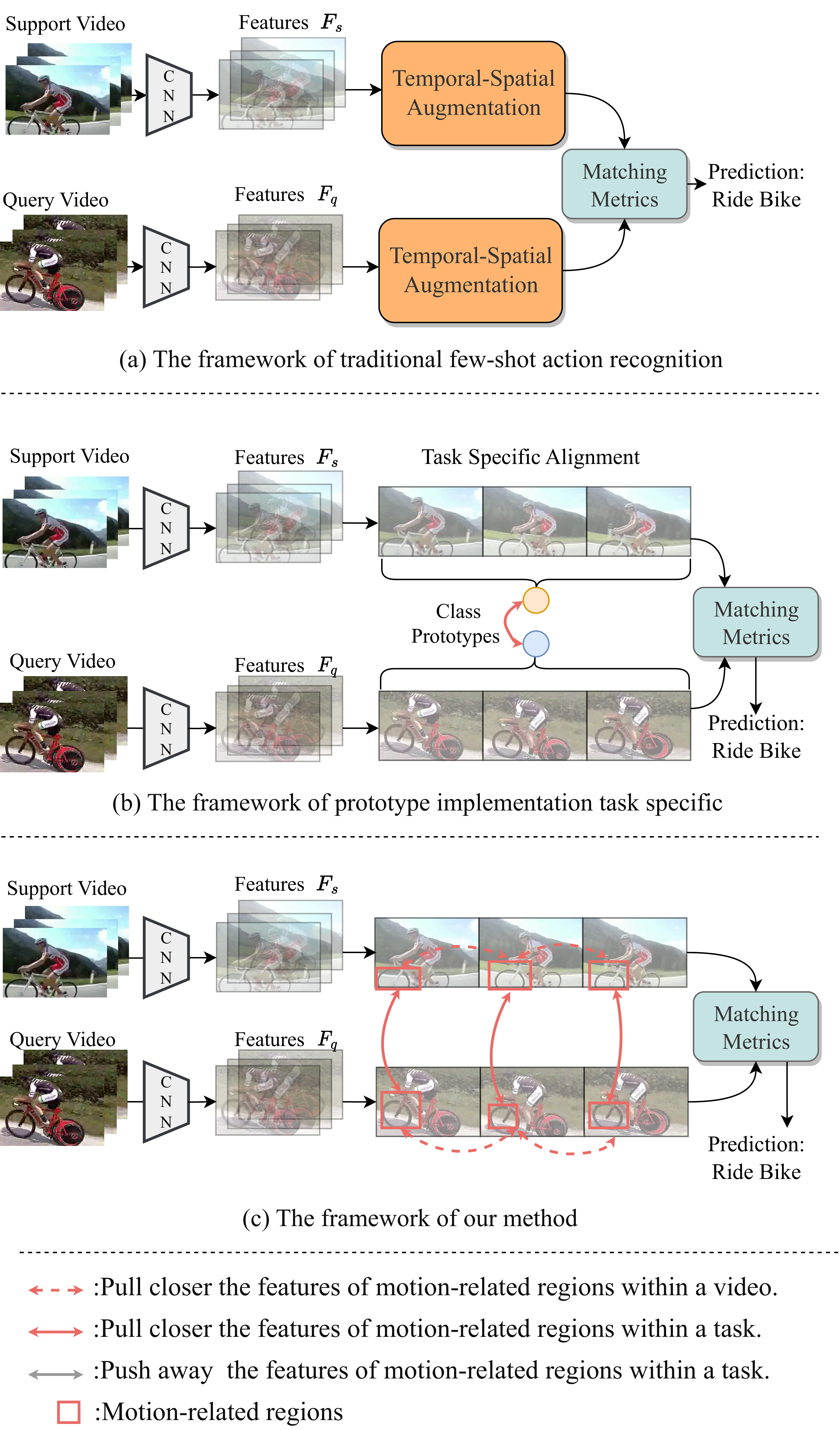}
	\caption{Comparison with previous methods. (a) Traditional methods design spatial and temporal alignment modules only to improve performance at the video level. (b) Recent works have focused on utilizing class prototypes to learn task-specific features, overlooking the spatio-temporal relationships between different videos at the task level. (c) Our method conducts spatio-temporal relation modeling for video-to-task learning, aiming to identify and correlate motion-related region features from the video level to the task level.}
	\label{fig:1}
\end{figure}


\section{Introduction}
\IEEEPARstart{T}{he} field of action recognition \cite{lin2019tsm, zhang2018learning, liu2022motion, simonyan2014two, zhang2018real} has achieved great success with the development of deep learning. However, strong performance in this field typically relies on a large number of labeled instances for training. To address the problem of data-deficient and resource-limited scenarios, few-shot learning \cite{finn2017model, cao2022learning, zhou2022hierarchical, das2019two} has become a primary focus of research in various fields, aiming to achieve satisfactory classification performance on unseen class data using only a few samples. While the majority of few-shot learning models for images \cite{li2024scformer} have achieved great success, significant challenges remain in extending them to videos due to the additional temporal dimension \cite{wang2024cross}. 

Similar to few-shot image recognition, existing few-shot action recognition methods \cite{cao2020few, thatipelli2022spatio, zhang2020few, zhu2018compound} match support and query videos using various class prototypes for recognition. However, designing an algorithm that fully incorporates spatio-temporal relationships into class prototypes remains a core challenge. To fully utilize spatio-temporal information, earlier works \cite{bishay2019tarn, cao2020few, li2022ta2n, wang2023molo} established various hand-designed spatio-temporal modeling algorithms to obtain more discriminative prototypes. Despite impressive progress, these methods often overlooked the spatio-temporal relationships across different videos at the task level. In data-scarce scenarios, fully considering task level features is beneficial for few-shot learning tasks \cite{lee2022task, wu2023bi}, as it can potentially increase inter-class variations while reducing intra-class variations.

Recent works (e.g., MTFAN~\cite{wu2022motion} and HyRSM~\cite{wang2022hybrid}) emphasize the importance of learning task-specific features. MTFAN proposes a motion encoder to learn global motion patterns and injects them into each video representation for task-specific feature learning. HyRSM introduces a hybrid relation module to generate task-specific features by capturing rich temporal information from different videos within an episode. Although MTFAN and HyRSM focus on enhancing task level features and utilize class prototypes for learning task-specific features, they still overlook the spatio-temporal relationships across different videos at the task level, especially in the spatial dimension, where these relationships provide rich information. However, conducting spatio-temporal relation modeling at the task level presents significant challenges due to the enormous computational and memory demands, as well as the difficulty in learning overly sparse spatio-temporal correspondences in few-shot scenarios. Therefore, it is highly meaningful to seek an efficient and effective spatio-temporal relation modeling approach at the task level for few-shot action recognition.

To achieve efficient and effective learning of task-specific spatio-temporal features without significant overheads and without sparse spatio-temporal correspondences that hinder the learning process, we propose a novel Dual Motion-Guided Attention Learning method (called DMGAL) for few-shot action recognition. As shown in Figure \ref{fig:1}, different from previous methods \cite{thatipelli2022spatio, li2022ta2n, wang2022hybrid, wu2022motion}, we conduct spatio-temporal relation modeling for video-to-task learning, aiming to identify and correlate motion-related region features from the video level to the task level. This process is similar to human brain reasoning \cite{martin2023seeing}: by first focusing on identifying the spatio-temporal relationships of actions within a single video and then extending to the spatio-temporal relationships of actions across different videos within the task, better performance can be achieved.

More specifically, we use carefully designed motion-guided attention to achieve video-to-task learning in an efficient and effective manner. This method consists of two key modules: Self Motion-Guided Attention module (S-MGA) and Cross Motion-Guided Attention module (C-MGA). Both modules are efficient in terms of parameters and computation. Additionally, due to their low parameter requirements and high adaptability, they can also serve as adapters within the adapter-tuning paradigm \cite{pan2022st} to optimize the model. As shown in Figure~\ref{fig:2}(a), we design the S-MGA module to focus on achieving spatio-temporal relation modeling at the video level. This module exploits features related to motion-related regions within a video, utilizing self motion-guided attention to guide enhancement of the most motion-related feature regions at the patch level. Finally, the S-MGA can achieve global context awareness of the entire video action evolution through a simple temporal MLP layer, providing a more comprehensive understanding of single video data and laying the groundwork for the subsequent learning of task-specific spatio-temporal features. Furthermore, based on the extracted video level spatio-temporal features, we extend S-MGA to a task level spatio-temporal relation modeling module, C-MGA. Different from S-MGA, the C-MGA module focuses on exploiting the correlation of motion-related region features across different videos within a task. It utilizes cross motion-guided attention to conduct frame-wise spatio-temporal relation modeling at the task level, pulling closer motion-related region features of the same class and pushing away motion-related region features of different classes in the embedding space. In this way, our method can gradually learn features related to the video level and the task level motion-related regions, allowing the model to construct class prototypes that fully incorporate spatio-temporal relationships from the video-specific level to the task-specific level.

We validate the efficiency and effectiveness of the proposed method under either fully fine-tuning \cite{perrett2021temporal} and adapter-tuning paradigms \cite{pei2023d}. The models developed using these paradigms are termed DMGAL-FT and DMGAL-Adapter, respectively. In DMGAL-FT, we follow previous methods \cite{perrett2021temporal, wu2022motion} by adopting the ResNet-50 \cite{he2016deep} pretrained on ImageNet \cite{deng2009imagenet} as our feature extractor and then constructing class prototypes through S-MGA and C-MGA that fully incorporate spatio-temporal relationships from video-specific to task-specific. We optimize the model through the fully fine-tuning paradigm. In DMGAL-Adapter, we simply plug the S-MGA and C-MGA as adapters into the pretrained CLIP-VIT \cite{radford2021learning} model and keep the model frozen. The adapter-tuning paradigm is a parameter-efficient technique \cite{xin2024parameter} that transfers the knowledge of foundational image models to achieve superior video understanding in data-deficient scenarios, thereby avoiding overfitting in few-shot learning caused by fully fine-tuning the pretrained models. By plugging in the S-MGA and C-MGA as adapters, the model can also learn class prototypes that fully incorporate spatio-temporal relationships from the video-specific level to the task-specific level within the adapter-tuning paradigm.

We extensively evaluate the performance of our DMGAL-FT and DMGAL-Adapter on five widely used benchmarks. The experimental results demonstrate that the proposed DMGAL-FT and DMGAL-Adapter significantly outperform several state-of-the-art methods by a substantial margin. In summary, the main contributions of this paper are as follows:
\begin{itemize}
\item We propose a novel Dual Motion-Guided Attention Learning method (called DMGAL) for few-shot action recognition, aiming to learn the spatio-temporal relationships from the video-specific level to the task-specific level.
\item We design S-MGA and C-MGA to efficiently and effectively identify and correlate motion-related region features from the video level to the task level, allowing the model to construct class prototypes that fully incorporate spatio-temporal relationships.
\item We develop two different models termed DMGAL-FT and DMGAL-Adapter, which are designed for fully fine-tuning and adapter-tuning paradigms, respectively.
\item We conduct extensive experiments on five few-shot action recognition benchmarks. The experimental results demonstrate the effectiveness and efficiency of the proposed DMGAL-FT and DMGAL-Adapter.  
\end{itemize}

\section{Related Work}
\subsection{Few-Shot Action Recognition}
The majority of existing methods on few-shot action recognition follow the metric learning paradigm, focusing primarily on modeling spatio-temporal relations or designing an effective matching strategy. OTAM \cite{cao2020few} designs a variant of the Dynamic Time Warping (DTW) algorithm to perform explicit temporal matching. TARN \cite{bishay2019tarn} utilizes attention mechanisms to perform temporal alignment at the video-segment level. CMN \cite{zhu2018compound} introduces a multi-saliency embedding algorithm that
encodes variable-length video sequences into fixed-size matrix representations for video matching. STRM \cite{thatipelli2022spatio} enhances local patch and global frame representations respectively to alleviate the spatio-temporal misalignment issues. TRX \cite{perrett2021temporal} matches each query sub-sequence with all sub-sequences in the support set and then aggregates these matching scores, allowing it to match actions performed at different speeds and in different parts of videos. TA2N \cite{li2022ta2n} devises a two-stage action alignment network to perform a spatial-temporal action alignment over videos jointly. MoLo \cite{wang2023molo} proposes a motion auto-decoder that learns to recover frame differences to explicitly extract motion representations. However, all the aforementioned works focus on individual video units or category units, missing the rich information between different classes within each task. While there is some research \cite{wu2022motion, wang2022hybrid} on task-specific learning, these methods still overlook the spatio-temporal relationships across different videos at the task level. In this work, we propose an efficient and effective spatio-temporal relation modeling approach at the task level for few-shot action recognition.

\subsection{Parameter-Efficient Fine-Tuning}
Parameter-Efficient Fine-Tuning (PEFT) \cite{xin2024parameter} aims to reduce the cost of fully fine-tuning large models by updating only a small subset of parameters and freezes the rest. Originally developed in NLP \cite{houlsby2019parameter}, adapter-tuning has been introduced to computer vision, allowing models to adapt efficiently to new tasks with minimal parameter updates while maintaining performance similar to fully fine-tuning.
ViT-Adapter \cite{chen2022vision} introduces adapter modules into the ViT \cite{dosovitskiy2020image} backbone, enabling efficient task-specific feature learning with minimal fine-tuning. ActionCLIP \cite{wang2021actionclip} incorporates temporal modeling into the frozen CLIP \cite{radford2021learning} backbone, capturing motion dynamics for effective video action recognition. CLIP-Adapter \cite{gao2024clip} adds a learnable adapter layer to the frozen CLIP backbone, allowing efficient adaptation to new tasks with minimal parameter updates. Med-SA \cite{wu2023medical} integrates adapter layers into the SAM \cite{kirillov2023segment} model, enabling fine-tuning for medical image segmentation tasks while preserving most of the pre-trained parameters. 
The aforementioned PEFT methods have developed various adapter modules designed for different tasks. To effectively combine adapter-tuning with few-shot action recognition \cite{pei2023d}, it is important to consider the characteristics of few-shot learning, where informative relationships between videos of different classes can help generate more discriminative features.
In this work, we propose a novel PEFT method that enhances the generalization ability of the pretrained model and equips it with the capacity for efficient spatio-temporal relation modeling at the task level.

\begin{figure*}
	\centering
	\includegraphics[width=\linewidth]{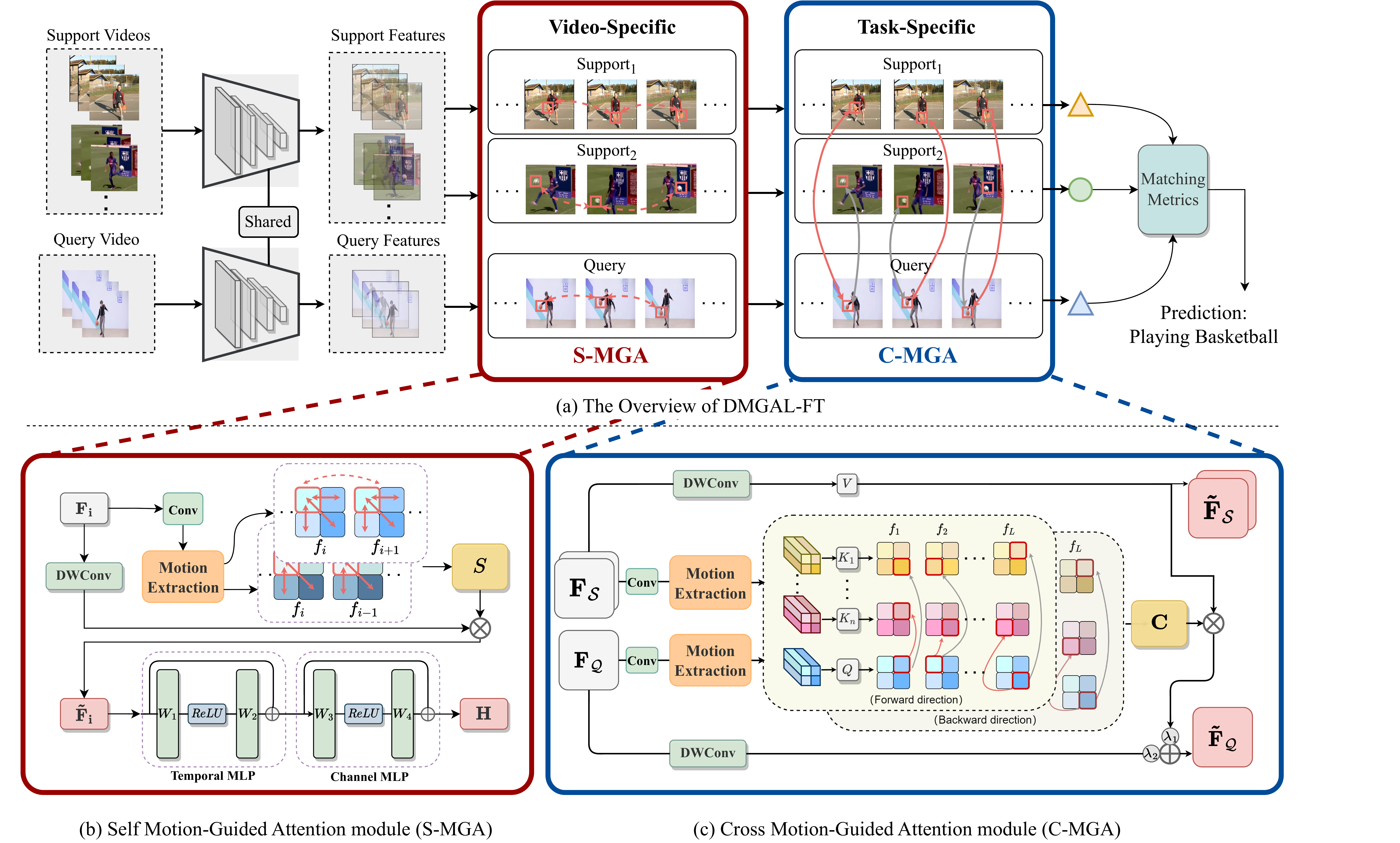}
	\caption{The overview of DMGAL-FT and details of our proposed MGA. (a) Overview of the DMGAL-FT model designed for the fully fine-tuning paradigm. DMGAL-FT uses S-MGA and C-MGA as additional modules to enable the model to sequentially learn spatio-temporal relationships at the video level and the task level, thereby achieving video-to-task learning via motion-guided attention within the fully fine-tuning paradigm. (b) Self Motion-Guided Attention module (S-MGA). S-MGA focuses on learning spatio-temporal relationships within a video, identifying and correlating motion-related region features in a video-specific manner. (c) Cross Motion-Guided Attention module (C-MGA). C-MGA focuses on learning spatio-temporal relationships within a task, identifying and correlating motion-related region features in a task-specific manner.}
	\label{fig:2}
\end{figure*}

\section{Proposed method}
In this section, we first outline the problem formulation of few-shot action recognition. Then, we introduce the key module MGA, providing a detailed explanation of how MGA achieves efficient and effective learning of video-specific to task-specific spatio-temporal features without significant overhead. Finally, we present two models developed with MGA: DMGAL-FT and DMGAL-Adapter, which are tailored for fully fine-tuning and adapter-tuning paradigms, respectively.

\subsection{Problem Formulation}
The task of few-shot action recognition aims to recognize novel classes using only a few labeled video samples. The task typically contains a training set $\mathcal{D}_{train}=\{(v_i,y_i),y_i \in \mathcal{C}_{train}\}$ for training the few-shot model and a test set $\mathcal{D}_{test}=\{(v_i,y_i),y_i \in \mathcal{C}_{test}\}$ for testing, where $\mathcal{C}_{train} \cap \mathcal{C}_{test}= \emptyset$. We adapt the episodic training paradigm following previous works \cite{thatipelli2022spatio, perrett2021temporal}, where each episode contains a support set $\mathcal{S}$ sampled from $N$ action classes with $K$ videos per class randomly selected from $\mathcal{D}_{train}$.
Subsequently, a query set $\mathcal{Q}$ samples are selected from the remaining videos of the $N$ action classes in $\mathcal{D}_{train}$.
For each episode, we expect to classify videos in the query set into one of the $N$ classes with the assistance  of the support set. 
For inference, we calculate the average accuracy of recognition performances over all episodes on the test set $\mathcal{D}_{test}$. 

\subsection{Motion-Guided Attention}
We achieve video-to-task learning via Motion-Guided Attention (MGA) for few-shot action recognition, aiming to learn the spatio-temporal relationships from the video-specific level to the task-specific level. MGA is composed of two key modules: Self Motion-Guided Attention module (S-MGA) and Cross Motion-Guided Attention module (C-MGA). S-MGA and C-MGA sequentially identify and correlate motion-related region features within videos and the task, thereby enabling the learning of video-specific to task-specific spatio-temporal relationships.

\subsubsection{Self Motion-Guided Attention}
S-MGA focuses on learning spatio-temporal relationships within a video, identifying and correlating motion-related region features in a video-specific manner. Specifically, S-MGA employs bidirectional and multi-scale motion features to learn a motion-related self-association score matrix, which explicitly represents the degree of relevance of the most motion-related region features within a video. Consequently, we can guide the correlation of motion-related region features in a video-specific manner.

As shown in Fig.~\ref{fig:2}(b), we extract motion features through a bidirectional multi-scale motion extraction module following BiMACL \cite{guo2024bi}. It is notable that before motion extraction, we compress the channels by a factor of $r_1$ using a $1 \times 1$ convolutional layer. Ablation experiments demonstrate that this approach enhances the efficiency and effectiveness of MGA in identifying and correlating motion-related regions. Subsequently, we calculate the aligned temporal differences between adjacent segments:
\begin{equation} 
\label{eq:1}
	\mathbf{C}(\mathbf{F}_i,\mathbf{F}_{i+1})=\mathbf{F}_i-\Phi_{2}(\mathbf{F}_{i+1}),
\end{equation}
where $\mathbf{C}(\mathbf{F}_i,\mathbf{F}_{i+1})$ represents the aligned temporal difference between frame $\mathbf{F}_i$ and frame $\mathbf{F}_{i+1}$. $\Phi_{2}(\cdot)$ represents a 2D depth-wise convolutional layer used for spatial smoothing to reduce the misalignment issue with low overhead. Similar to BiMACL, we extract motion features through a multi-scale module:
\begin{align}
& \begin{aligned}
\label{eq:2}
	\mathbf{M}(\mathbf{F}_i,\mathbf{F}_{i+1})=\text{Conv}\left(\frac{1}{N} \sum \limits _{j=1} ^{N} \text{CNN}_j(\mathbf{C}(\mathbf{F}_i,\mathbf{F}_{i+1}))\right),
\end{aligned} \\
& \begin{aligned}
\label{eq:3}
	\mathbf{M}(\mathbf{F}_{i+1},\mathbf{F}_{i})=\text{Conv}\left(\frac{1}{N} \sum \limits _{j=1} ^{N} \text{CNN}_j(\mathbf{C}(\mathbf{F}_{i+1},\mathbf{F}_{i}))\right),
\end{aligned}
\end{align}
where $N=3$ in practice corresponds to the number of $\text{CNN}_j$ at different spatial scales, each extracting motion information from distinct receptive fields. We follow the implementation of the three branches: $a)$ a short connection; $b)$ a $3 \times 3$ convolution; $c)$ an average pooling, a $3 \times 3$ convolution, and a bilinear upsampling. $\text{Conv}$ is a channel-wise convolution used to aggregate multi-scale motion features. The bidirectional and multi-scale motion features contain rich motion cues between frames. Therefore, we suggest utilizing them to capture accurate motion-related region correlations between frames in each video.

Subsequently, we utilize self-attention with bidirectional motion features to obtain the motion-related self-association score matrix $\mathbf{S}_i$ of frame-level features $\mathbf{F}_i$ in order to explicitly convey the degree of relevance of the most motion-related regions in $\mathbf{F}_i$ with adjacent frames at the patch level:
\begin{align}
	& \begin{aligned}
	\mathbf{S}^B_i = \frac{\mathbf{M}(\mathbf{F}_i, \mathbf{F}_{i+1}) \mathbf{M}(\mathbf{F}_i, \mathbf{F}_{i+1})^\top}{\sqrt{D/r_1}},
	\end{aligned} \\
	& \begin{aligned}
	\mathbf{S}^F_i = \frac{\mathbf{M}(\mathbf{F}_{i+1}, \mathbf{F}_i) \mathbf{M}(\mathbf{F}_{i+1}, \mathbf{F}_i)^\top}{\sqrt{D/r_1}},
	\end{aligned} \\
 	& \begin{aligned}
	\mathbf{S}_i = \eta \left(\mathbf{S}^B_i + \mathbf{S}^F_i \right),
        \label{eq:3}
	\end{aligned}
\end{align}
where $\eta(\cdot)$ denotes the softmax function, and $D$ is the dimension of frame-level features. $\mathbf{S}^B_i$ and $\mathbf{S}^F_i$ represent the backward and forward self-association score matrices, respectively. A higher score signifies stronger relevance of motion-related regions for a specific path. Therefore, through the motion-related self-association score matrix $\mathbf{S}_i$, we can guide the correlation of motion-related region features in a manner specific to each video. We compute video-specific motion-related region highlight features $\mathbf{\tilde{F}}_i$ guided by $\mathbf{S}_i$:
\begin{equation} 
\label{eq:4}
	\mathbf{\tilde{F}}_i=\lambda \cdot \mathbf{S}_i \Phi_{3}(\mathbf{F})_i + \mathbf{F}_i,\quad i=1...L
\end{equation}
where $\lambda$ is the learnable parameter, and $L$ is the number of sampled frames in a video. $\Phi_{3}(\cdot)$ is the 3D depth-wise convolution utilizing $3 \times 1 \times 1$ convolutional kernels, which can enhance the modeling ability of motion-related regions with little spatial variation in the temporal dimension. Therefore, the module also has strong awareness of regions that are less sensitive to motion.

Although Eq.~(\ref{eq:3}) and Eq.~(\ref{eq:4}) can effectively and efficiently perform video-specific motion-related region correlations, they primarily rely on motion information from adjacent frames, resulting in a limited receptive field in the temporal dimension. To achieve a more global motion-related region correlation within a video, we use the MLP-mixer \cite{tolstikhin2021mlp} to effectively enlarge the receptive field of patches in the feature space along the temporal dimension. Given a single video data $\mathbf{F}\in \mathbb{R}^{L \times D \times H \times W}$, we can describe the process using the following formula:
\begin{align}
\label{eq:5}
	\mathbf{G}_{i,*,*,*} &=\mathbf{W}_2\sigma(\mathbf{W}_1\mathbf{\tilde{F}}_{i,*,*,*})+\mathbf{\tilde{F}}_{i,*,*,*},\quad i=1...L
\end{align} 
\begin{align}
\label{eq:6}
	\mathbf{H}_{*,j,*,*} &= \mathbf{W}_4\sigma(\mathbf{W}_3 \mathbf{G}_{*,j,*,*})+\mathbf{G}_{*,j,*,*},\quad j=1...D 
\end{align} 
where '*' represents a certain feature dimension, $\sigma(\cdot)$ is the ReLU non-linearity, $\mathbf{W}_1,\mathbf{W}_2 \in \mathbb{R}^{L \times L}$ are the weights of the temporal MLP, and $\mathbf{W}_3,\mathbf{W}_4 \in \mathbb{R}^{D \times D}$ are the weights of the channel MLP. By effectively enlarging the receptive field in the feature space along the temporal dimension, it enhances temporal global context awareness and provides a more comprehensive understanding of video data, thereby implicitly correlating motion-related region features between different frames within the entire video. We also verified this in visualization experiments.

\subsubsection{Cross Motion-Guided Attention}
C-MGA focuses on learning spatio-temporal relationships within a task, identifying and correlating motion-related region features in a task-specific manner. Specifically, C-MGA employs bidirectional and multi-scale motion features from all videos in the task to learn multiple motion-related cross-association score matrices, which explicitly represent the degree of relevance of the most motion-related region features within the task. Consequently, we can guide the correlation of motion-related region features in a task-specific manner.

However, conducting spatio-temporal relation modeling at the task level presents two significant challenges: 
\(i) \) enormous computational and memory demands; 
\(ii) \) difficulty in learning overly sparse spatio-temporal correspondences in few-shot scenarios.
In our task-specific method, we address these issues through frame-wise cross motion-guided attention.

C-MGA employs convolution to compress the dimension by a factor of $r_2$, and then extracts the bidirectional and multi-scale motion features of the entire task using the same approach as S-MGA. The difference is that C-MGA extracts the motion features for the entire task. By compressing the dimensions, we can significantly reduce the computational and memory demands required for task-specific spatio-temporal relation modeling. Experimental results also indicate that this approach can even enhance performance.

Subsequently, for a query video comprising $L$ frames, C-MGA uses frame-wise approach to calculate $L$ motion-related cross-association score matrices, indicating the relevance of motion-related region features between query video and all support videos within the entire task at patch level. It is worth noting that by adopting frame-wise approach, C-MGA can further reduce the computational load by $L$ times and mitigate the difficulty of learning overly sparse spatio-temporal correspondences in few-shot scenarios, thereby achieving better results consistent with our experimental results.

\begin{figure*}
	\centering
	\includegraphics[width=\linewidth]{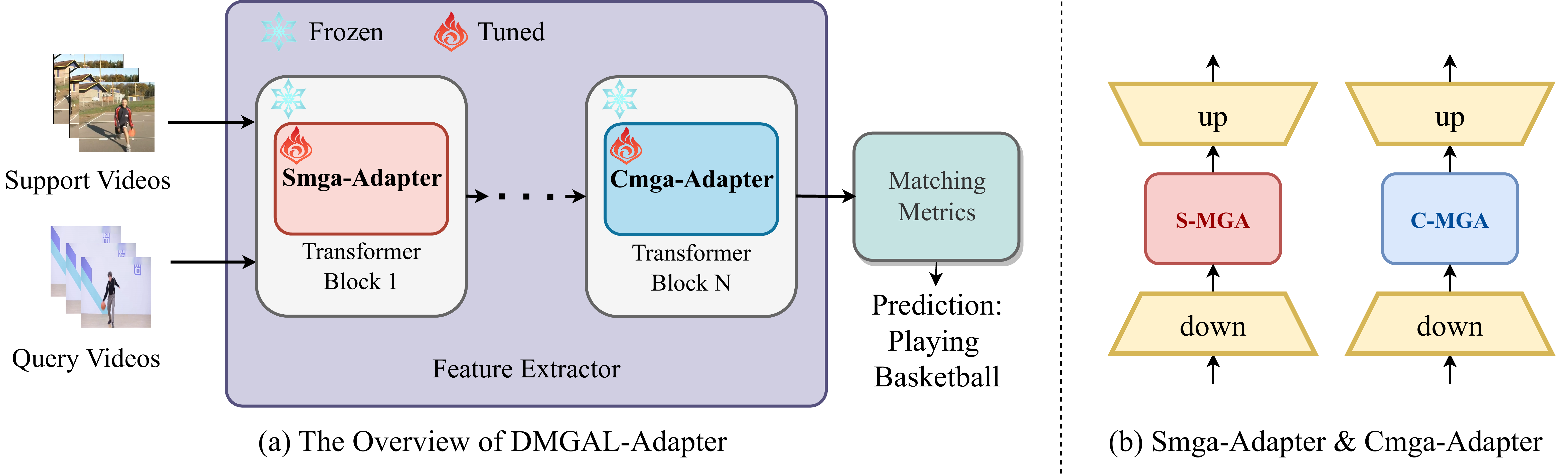}
	\caption{The overview of DMGAL-Adapter (a) Overview of the DMGAL-Adapter model designed for the adapter-tuning paradigm. DMGAL-Adapter selectively plugs the Smga-Adapter into all the early layers of a pre-trained model, reserving the final layer for the Cmga-Adapter. (b) Details of the Smga-Adapter and Cmga-Adapter, which are simplified versions of S-MGA and C-MGA, respectively.}
	\label{fig:3}
\end{figure*}

Specifically, in a task that includes a query set $\mathcal{Q}$ and a support set $\mathcal{S}$, for each query video, C-MGA generates $L$ unique motion-related cross-association score matrices, each corresponding to one of the $L$ frames in the video. The $i^{th}$ motion-related cross-association score matrix for the $i^{th}(i \in [1,L])$ frame in the $m^{th}(m \in [1,\left|\mathcal{Q}\right|])$ video within the query set $\mathcal{Q}$ is computed as follows:
\begin{align}
	& \begin{aligned}
	\mathbf{C}^B_{\mathcal{Q},m,i} = \frac{\mathbf{M}_{\mathcal{Q},m}(\mathbf{F}_i, \mathbf{F}_{i+1})\mathbf{M}_{\mathcal{S}}(\mathbf{F}_i, \mathbf{F}_{i+1})^T}{\sqrt{D/r_2}},
	\end{aligned} \\
	& \begin{aligned}
	\mathbf{C}^F_{\mathcal{Q},m,i} = \frac{\mathbf{M}_{\mathcal{Q},m}(\mathbf{F}_{i+1}, \mathbf{F}_i) \mathbf{M}_{\mathcal{S}}(\mathbf{F}_{i+1}, \mathbf{F}_i)^T}{\sqrt{D/r_2}},
	\end{aligned} \\
 	& \begin{aligned}
	\mathbf{C}_{\mathcal{Q},m,i} = \eta \left(\mathbf{C}^B_{\mathcal{Q},m,i} + \mathbf{C}^F_{\mathcal{Q},m,i} \right),
        \label{eq:7}
	\end{aligned}
\end{align}
where $\eta(\cdot)$ denotes the softmax function, $\mathbf{M}_{\mathcal{Q},m}(\cdot, \cdot)$ represents the motion features of the $m^{th}$ video within the query set $\mathcal{Q}$,
$\mathbf{M}_{\mathcal{S}}(\cdot, \cdot)$ represents the motion features of the entire support set $\mathcal{S}$, $\mathbf{C}^B_{\mathcal{Q},m,i}$ and $\mathbf{C}^F_{\mathcal{Q},m,i}$ represent the backward and forward cross-association score matrices, respectively. $\mathbf{M}_{\mathcal{Q},m} \in \mathbb{R}^{L \times HW \times D/r_2}$, $\mathbf{M}_{\mathcal{S}} \in \mathbb{R}^{L \times NKHW \times D/r_2}$, hence $\mathbf{C}^B_{\mathcal{Q},m,i}$, $\mathbf{C}^F_{\mathcal{Q},m,i}$, $\mathbf{C}_{\mathcal{Q},m,i} \in \mathbb{R}^{HW \times NKHW}$. $HW$ represents the spatial points of a frame. We compute the motion-related cross-association score matrix between the spatial points of the $i^{th}$ frame in the $m^{th}$ video of the query set and all spatial points of the $i^{th}$ frame across the entire support set. A higher score signifies stronger relevance of motion-related regions to the task. Therefore, through the motion-related cross-association score matrix $\mathbf{C}_{\mathcal{Q},m,i}$, we can guide the correlation of motion-related region features in a task-specific manner. Succinctly, the enhanced features of the query set and support set are defined as:
\begin{align}
	& \begin{aligned}
	   \mathbf{\tilde{F}}_{\mathcal{Q}, m,i} = \lambda_{1} \cdot \mathbf{C}_{\mathcal{Q},m,i} \Phi_{3}(\mathbf{F}_{\mathcal{S}})_i+ \lambda_{2} \cdot \Phi_{3}(\mathbf{F}_{\mathcal{Q},m})_i + \mathbf{F}_{\mathcal{Q},m,i},
	\end{aligned} \\
	& \begin{aligned}
		\mathbf{\tilde{F}}_{\mathcal{S},n,i} = \lambda_{2} \cdot \Phi_{3}(\mathbf{F}_{\mathcal{S},n})_i + \mathbf{F}_{\mathcal{S},n,i},
	\end{aligned}
\end{align}
where $\lambda_1$ and $\lambda_2$ are the learnable parameters. The variable $n$ ranges from 1 to $N \times K$, representing the $n^{th}$ video in the support set. $\Phi_{3}(\cdot)$ is a 3D depth-wise convolutional layer consistent with S-MGA. $\mathbf{F}_{\mathcal{Q},m,i}, \mathbf{F}_{\mathcal{S},n,i}, \mathbf{\tilde{F}}_{\mathcal{Q},m,i}, \mathbf{\tilde{F}}_{\mathcal{S},n,i} \in \mathbb{R}^{D \times H \times W}$ represent frame-level features. This correlation enables direct learning of motion-related relationships within a task, promoting higher scores for motion-related regions of the same class and lower scores for motion-related regions of different classes. Therefore, it increases inter-class variations while reducing intra-class variations, which is consistent with our visualization experiments.

By sequentially identifying and correlating motion-related region features within videos and the task through S-MGA and C-MGA, our method can gradually learn both video-level and task-level motion-related features. This allows the model to construct class prototypes that fully incorporate spatio-temporal relationships from the video-specific level to the task-specific level.

\subsection{Models: DMGAL-FT and DMGAL-Adapter}
To validate the efficiency and effectiveness of our proposed Dual Motion-Guided Attention Learning method (called DMGAL), we developed two models for training using the fully fine-tuning and the adapter-tuning paradigms, respectively: DMGAL-FT and DMGAL-Adapter. Both models are designed based on S-MGA and C-MGA, enabling them to gradually learn the correlation of motion-related regions from the video-specific level to the task-specific level. The details of DMGAL-FT and DMGAL-Adapter are described as follows:

\subsubsection{DMGAL-FT}
DMGAL-FT is specifically designed for the fully fine-tuning paradigm following traditional few-shot action recognition methods. In this paradigm, models require an additional module to learn the spatio-temporal relationships as a post-process after feature learning by the pre-trained backbone. During training, the parameters of the entire model are optimized.

The overview of DMGAL-FT is shown in Fig. \ref{fig:2}(a). We use MGA as additional modules to enable the model to sequentially learn spatio-temporal relationships at the video level and the task level, making video-to-task learning via motion-guided attention within the fully fine-tuning paradigm.

\begin{table*}[t]
    \centering
    \caption{More statistics of the datasets and commonly used experimental settings for few-shot action recognition.}
    \label{tab:1}
    \scalebox{1.15}{
    \begin{tabular}{
        >{\raggedright\arraybackslash}m{2.0cm} 
        >{\raggedright\arraybackslash}m{1.8cm} 
        >{\centering\arraybackslash}m{1.5cm} 
        >{\centering\arraybackslash}m{1.0cm}
        >{\centering\arraybackslash}m{1.0cm} 
        >{\centering\arraybackslash}m{1.0cm}
        >{\centering\arraybackslash}m{1.0cm}
        >{\centering\arraybackslash}m{1.0cm} 
        >{\centering\arraybackslash}m{1.0cm}}
        \toprule
        \multirow{2}{*}{\textbf{Datasets}} & \multirow{2}{*}{\textbf{Split reference}} & \multirow{2}{*}{\textbf{Modality}} 
        & \multicolumn{3}{c}{\textbf{Classes}} & \multicolumn{3}{c}{\textbf{Videos}} \\
        \cmidrule(lr){4-6} \cmidrule(lr){7-9}
        & & & \textbf{Train} & \textbf{ Val} & \textbf{Test} & \textbf{Train} & \textbf{ Val} & \textbf{Test} \\
        \midrule
        SSv2-Full \cite{goyal2017something}  & OTAM \cite{cao2020few} & Temporal & 64 & 12 & 24 & 77500 & 1925 & 2854 \\
        SSv2-Small \cite{goyal2017something} & CMN \cite{zhu2018compound}  & Temporal & 64 & 12 & 24 & 6400  & 1200 & 2400 \\
        Kinetics \cite{carreira2017quo}   & CMN \cite{zhu2018compound}  & Spatial  & 64 & 12 & 24 & 6400  & 1200 & 2400 \\
        HMDB51 \cite{kuehne2011hmdb}     & ARN \cite{zhang2020few}  & Spatial  & 31 & 10 & 10 & 4280  & 1194 & 1292 \\
        UCF101 \cite{soomro2012ucf101}     & ARN \cite{zhang2020few}  & Spatial  & 70 & 10 & 21 & 9154  & 1421 & 2745 \\
        \bottomrule
    \end{tabular}}
\end{table*}

\subsubsection{DMGAL-Adapter}
DMGAL-Adapter is designed following the adapter-tuning paradigm. Inspired by ST-Adapter \cite{pan2022st}, which first applied the adapter-tuning paradigm to learn spatio-temporal features while maintaining a lightweight structure, recent works \cite{pei2023d} have attempted to use the same paradigm for few-shot action recognition. In this paradigm, a lightweight module (named Adapter) is selectively plugged into a pre-trained model to learn spatio-temporal features. During training, only the parameters of the adapter are optimized. The ST-Adapter can be expressed as:
\begin{equation}
\text{ST-Adapter}(\mathbf{X}) = \mathbf{X} +  \text{DWConv3D}(\mathbf{X}\mathbf{W}_{\text{down}}) \mathbf{W}_{\text{up}},
\end{equation}
where DWConv3D$(\cdot)$ denotes the depth-wise 3D convolution for spatio-temporal relationship modeling. $\mathbf{W}_{\text{down}} \in \mathbb{R}^{D \times D'}$ and $\mathbf{W}_{\text{up}} \in \mathbb{R}^{D' \times D}$ are two linear weights for downsampling and upsampling, respectively. $\mathbf{X}$ represents the abstract input. Although the ST-Adapter and its subsequent works \cite{pei2023d} using the adapter-tuning paradigm have continuously achieved satisfactory results, they primarily focus on image-to-video transfer and fail to consider task-level spatio-temporal relationship modeling. In DMGAL-Adapter, we simply plug the S-MGA and C-MGA as adapters to perform adapter-tuning. Their structure is illustrated in Figure~\ref{fig:3}(b) and can be expressed as follows:
\begin{equation}
\text{Smga-Adapter}(\mathbf{X}) = \mathbf{X} +  \text{S-MGA}(\mathbf{X}\mathbf{W}_{\text{down}}) \mathbf{W}_{\text{up}},
\end{equation}
\begin{equation}
\text{Cmga-Adapter}(\mathbf{X}) = \mathbf{X} +  \text{C-MGA}(\mathbf{X}\mathbf{W}_{\text{down}}) \mathbf{W}_{\text{up}}.
\end{equation}

It is notable that the Smga-Adapter takes a video as input, while the Cmga-Adapter takes a task as input. The overview of DMGAL-Adapter is shown in Fig.~\ref{fig:3}(a). We selectively plug the Smga-Adapter into all the early layers of a pre-trained model, reserving the final layer for the Cmga-Adapter. This design enables the model to sequentially learn spatio-temporal relationships at the video level and the task level, making video-to-task learning via motion-guided attention within the adapter-tuning paradigm.

Finally, both DMGAL-FT and DMGAL-Adapter can be combined with various classical matching metrics for few-shot action classification. Experimental results have demonstrated their effectiveness.

\section{Experiments}
In this section, we first introduce the datasets used and provide the implementation details of the proposed method. After that, we describe the experimental results obtained by our DMGAL on both the fully fine-tuning (DMGAL-FT) and the adapter-tuning (DMGAL-Adapter) paradigms. Finally, we perform extensive ablation studies and present visualization results to further evaluate the effectiveness of the proposed DMGAL.

\subsection{Datasets}
We conduct experiments on five challenging few-shot action recognition datasets, including the temporal datasets SSv2-Full \cite{goyal2017something} and SSv2-Small \cite{goyal2017something}, as well as the spatial datasets Kinetics \cite{carreira2017quo}, HMDB51 \cite{kuehne2011hmdb}, and UCF101 \cite{soomro2012ucf101}. 

For SSv2-Full and SSv2-Small, we follow the splits provided in OTAM \cite{cao2020few} and CMN \cite{zhu2018compound}, where 64 classes were randomly selected for the training set and 24 classes were designated for the test set. SSv2-Full utilized all samples from each category in the original dataset \cite{goyal2017something}, while SSv2-Small only select 100 samples from each category. A similar split, as in CMN \cite{zhu2018compound}, with 64 classes for the training set and 24 classes for the testing set is used for Kinetics. For HMDB51 and UCF101, we follow the settings of ARN \cite{zhang2020few}, where 31 classes are selected for training and 10 classes for testing from HMDB51, while the 101 classes in UCF101 are divided into 70 for training and 21 for testing. More statistics of the datasets and commonly used experimental settings are listed in Table~\ref{tab:1}.

\subsection{Implementation details}
To ensure a fair comparison with previous methods \cite{perrett2021temporal, thatipelli2022spatio}, we follow the TSN \cite{wang2016temporal} setting to randomly sample 8 frames uniformly for each video, i.e., $L=8$. We then resize each frame to $256\times 256$ and crop it to $224\times 224$. Additionally, we employ common data augmentations such as random cropping, horizontal flipping, and color jitter during model training. Since the labels in the Something-Something V2 dataset include assumptions about left and right directions (e.g., pulling something from left to right or from right to left), we do not use horizontal flips for this dataset.

In DMGAL-FT, we use ResNet-50 \cite{he2016deep} pretrained on ImageNet \cite{deng2009imagenet} as our feature extractor, removing the last two layers to retain spatial motion information. Both factors $r_1$ in S-MGA and $r_2$ in C-MGA are set to 8. In DMGAL-Adapter, we choose CLIP-ViT-B \cite{radford2021learning} as the pretrained model to conduct adapter-tuning following previous work for a fair comparison. Both factor $r_1$ in Smga-Adapter and $r_2$ in Cmga-Adapter are set to 4. We evaluate our DMGAL using OTAM \cite{cao2020few}, Bi-MHM \cite{wang2022hybrid}, SCA \cite{zhang2023importance}, and TRX \cite{perrett2021temporal} as matching metrics for comparison with other methods to demonstrate the effectiveness and robustness of DMGAL. 

We employ four NVIDIA RTX3090 GPUs for training. We adopt Adam optimizer to optimize our model end-to-end, following the episodic training paradigm \cite{vinyals2016matching, snell2017prototypical}. The learning rate and other training parameters are configured according to Molo \cite{wang2023molo} and CLIP-FSAR \cite{wang2024clip}. During the testing stage, we randomly sample 10,000 episodes from the test set and report the average classification accuracy.

\begin{table*}[t]
	\centering
	\caption{Comparison with state-of-the-art methods on the SSv2-Small and SSv2-Full datasets using DMGAL-FT. '-' stands for the result is not available in published works. The best results are bolded in black and the underline represents the second best result. '*' stands for the results of our implementation. We highlight that: 1) 'X' in DMGAL-FT (X) represents one of the four matching metrics we used. Its results are marked in gray, along with the improvement value compared to using 'X' alone. 2) SCA is a novel matching metric proposed by SA-CT \cite{zhang2023importance} to align the spatial objects in the query and support set.}
	\label{tab:2}
    \scalebox{1.26}{
	\begin{tabular}{lcllllll}
	\toprule
	\multirow{2}[2]{*}{\textbf{Method}} & \multirow{2}[2]{*}{\textbf{Reference}} & \multicolumn{3}{c}{\textbf{SSv2-Small}} & \multicolumn{3}{c}{\textbf{SSv2-Full}} \\
    \cmidrule(lr){3-5} \cmidrule(lr){6-8}
	& & \textbf{1-shot} & \textbf{3-shot} & \textbf{5-shot} & \textbf{1-shot} & \textbf{3-shot} & \textbf{5-shot}\\
	\midrule
	HCL \cite{zheng2022few}   & ECCV'22  & 38.7  & 49.1  & 55.4  & 47.3  & 59.0  & 64.9 \\
	STRM \cite{thatipelli2022spatio}   & CVPR'22  & 37.1$^*$  & 49.2$^*$  & 55.3 & 43.1$^*$  & 59.1$^*$  & 68.1\\
    MPRE \cite{liu2022multidimensional} & TCSVT'22 & - & - & - & 42.1  & -  & 58.4 \\
    MTFAN \cite{wu2022motion}  & CVPR'22 & - & - & - & 45.7  & -  & 60.4 \\
    HyRSM \cite{wang2022hybrid} & CVPR'22 & 40.6 & 52.3 & 46.1 & 54.3 & 65.1 & 69.0 \\
    MoLo  \cite{wang2023molo}  & CVPR'23 & \underline{42.7}  & 52.9  & 56.4  & \underline{56.6}  & \underline{67.0}  & \underline{70.6}\\
    SA-CT \cite{zhang2023importance}  & ACM MM’23 & - & - & - & 48.9 & - & 69.1\\
    BiMACL \cite{guo2024bi} & ICASSP'24 & 38.6 & - & 61.1  & 45.6 & - & 70.2 \\
    huang et al. \cite{huang_ijcv} & IJCV'24 & 42.6 & - & \underline{61.8}  & 52.3 & - & 67.1\\
    CCLN \cite{wang2024cross} & TIP'24  & - & - & - & 46.0 & - & 61.3\\
	\midrule
    OTAM  \cite{cao2020few}  & CVPR’20 & 36.4 & - & 48.0  & 42.8 & - & 52.3\\
    \rowcolor{gray!20}
    DMGAL-FT  (OTAM) & -  & 40.9$_{\text{+}4.5}$ & -  & 58.1$_{\text{+}10.1}$  & 53.9$_{\text{+}11.1}$  & -  & 68.6$_{\text{+}16.3}$\\
    SCA \cite{zhang2023importance}   & ACM MM’23  & 38.8$^*$ & - & 58.0$^*$ & 41.7$^*$ & - & 55.7$^*$ \\
    \rowcolor{gray!20}
	DMGAL-FT (SCA) & -  & 42.2$_{\text{+}3.4}$ & - & 60.8$_{\text{+}2.8}$  & 55.3$_{\text{+}13.6}$ & - & 70.0$_{\text{+}14.3}$\\
    Bi-MHM  \cite{wang2022hybrid} & CVPR'22 & 38.0 & 47.6 & 47.9  & 44.6 & 53.1 & 56.0\\
    \rowcolor{gray!20}
    DMGAL-FT (Bi-MHM)  & - & \bfseries{42.8}$_{\text{+}4.8}$ & \underline{54.7}$_{\text{+}7.1}$ & 59.0$_{\text{+}10.1}$  & \bfseries{56.7}$_{\text{+}12.1}$ & 66.4$_{\text{+}14.3}$ & 69.5$_{\text{+}13.5}$\\
    TRX  \cite{perrett2021temporal}   & CVPR'21  & 36.0 & 51.9  & 59.1  & 42.0  & 57.6  & 64.6\\
    \rowcolor{gray!20}
    DMGAL-FT (TRX) & - & 41.2$_{\text{+}5.2}$ & \bfseries{56.3}$_{\text{+}4.4}$  & \bfseries{61.8}$_{\text{+}2.7}$ & 55.5$_{\text{+}13.5}$ & \bfseries{68.1}$_{\text{+}10.5}$  & \bfseries{71.1}$_{\text{+}6.5}$\\
    \bottomrule 
    \end{tabular}}
\end{table*}

\begin{table*}[t]
	\centering
	\caption{Comparison with state-of-the-art methods on the UCF101, HMDB51 and Kinetics datasets using DMGAL-FT. '-' stands for the result is not available in published works. The best results are bolded in black and the underline represents the second best result. '*' stands for the results of our implementation. We highlight that: 1) 'X' in DMGAL-FT (X) represents one of the four matching metrics we used. Its results are marked in gray, along with the improvement value compared to using 'X' alone. 2) SCA is a novel matching metric proposed by SA-CT \cite{zhang2023importance} to align the spatial objects in the query and support set.}
	\label{tab:3}
    \scalebox{1.3}{
		\begin{tabular}{lcllllllllll}
			\toprule
			\multirow{2}[2]{*}{\textbf{Method}} & \multirow{2}[2]{*}{\textbf{Reference}} & \multicolumn{2}{c}{\textbf{UCF101}} & \multicolumn{2}{c}{\textbf{HMDB51}} & \multicolumn{2}{c}{\textbf{Kinetics}} \\
            \cmidrule(lr){3-4} \cmidrule(lr){5-6} \cmidrule(lr){7-8}
			& & \textbf{1-shot} & \textbf{5-shot} & \textbf{1-shot} & \textbf{5-shot} & \textbf{1-shot} & \textbf{5-shot}\\
			\midrule
			HCL \cite{zheng2022few}     & ECCV'22  & 82.5  & 93.9  & 59.1  & 76.3  & 73.7  & 85.8 \\
			STRM \cite{thatipelli2022spatio}   & CVPR'22 & 80.5$^*$  & 96.9  & 52.3$^*$  & 77.3  & 62.9$^*$  & 86.7 \\
            MPRE \cite{liu2022multidimensional} & TCSVT'22 & 82.0  & 96.4  & 57.3    & 76.8  & 70.2  & 85.3\\
		MTFAN \cite{wu2022motion} & CVPR'22 & 84.8  & 95.1  & 59.0    & 74.6  & 74.6  & 87.4\\
            HyRSM  \cite{wang2022hybrid} & CVPR'22 & 83.9  & 94.7 & 60.3 & 76.0 & 73.7 & 86.1 \\
		MoLo  \cite{wang2023molo}  & CVPR'23 & 86.0  & 95.5  & 60.8  & 77.4  & 74.0  & 85.6\\
            SA-CT \cite{zhang2023importance} & ACM MM’23 & 85.4 &   96.4   & 60.4 & 78.3 & 71.9 & 87.1\\
            BiMACL \cite{guo2024bi} & ICASSP'24 & 83.9  & 97.2  & 57.0  & 78.4  & 68.1  & 87.6\\
            huang et al. \cite{huang_ijcv} & IJCV'24 & 74.9  & 92.5  & 61.6  & 77.5  & 74.0  & 86.9\\
            CCLN \cite{wang2024cross} & TIP'24 & 86.9  & 96.1  & \bfseries{65.1}  & \underline{78.8}  & \bfseries{75.8}  & 87.5\\
			\midrule
            OTAM \cite{cao2020few}  & CVPR’20 & 79.9  & 88.9  & 54.5  & 68.0    & 73.0    & 85.8\\
            \rowcolor{gray!20}
			DMGAL-FT  (OTAM) & -  & 87.8$_{\text{+}7.9}$   & 95.9$_{\text{+}7.0}$  & 63.2$_{\text{+}8.7}$  & 76.4$_{\text{+}8.4}$  & 74.2$_{\text{+}1.2}$  & 86.4$_{\text{+}0.6}$\\
            SCA \cite{zhang2023importance}  & ACM MM’23  & 85.2* &    96.0*   & 58.7* & 77.4* & 72.0* & 84.7*\\
            \rowcolor{gray!20}
			DMGAL-FT (SCA) & -  & \underline{87.4}$_{\text{+}2.2}$  & \underline{97.2}$_{\text{+}1.2}$  & 61.5$_{\text{+}2.8}$  & 78.6$_{\text{+}1.2}$  & 73.8$_{\text{+}1.8}$  & \underline{87.6}$_{\text{+}2.9}$ \\
            Bi-MHM \cite{wang2022hybrid} & CVPR'22 & 81.7  & 89.3  & 58.3  & 69.0    & 72.3  & 84.5\\
            \rowcolor{gray!20}
			DMGAL-FT (Bi-MHM)  & - & \bfseries{88.0}$_{\text{+}6.3}$ & 96.3$_{\text{+}7.0}$  & \underline{63.5}$_{\text{+}5.2}$ & 76.7$_{\text{+}7.7}$  & \underline{75.0}$_{\text{+}2.7}$ & 85.7$_{\text{+}1.2}$\\
            TRX \cite{perrett2021temporal}   & CVPR'21  & 78.2  & 96.1  & 53.1  & 75.6  & 63.6  & 85.9\\
            \rowcolor{gray!20}
			DMGAL-FT (TRX) & - & 87.1$_{\text{+}8.9}$  & \bfseries{97.4}$_{\text{+}1.3}$ & 60.2$_{\text{+}7.1}$  & \bfseries{79.3}$_{\text{+}3.7}$ & 72.5$_{\text{+}8.9}$  & \bfseries{88.0}$_{\text{+}2.1}$ \\
			\bottomrule
   
		\end{tabular}}
\end{table*}%

\begin{table*}[t]
	\centering
	\caption{Comparison with state-of-the-art methods on four datasets using DMGAL-Adapter. '-' stands for the result is not available in published works. The best results are bolded in black and the underline represents the second best result. '*' stands for the results of our implementation. 'X' in DMGAL-Adapter (X) represents the alignment metric we used. '$\dag$' and '$\ddag$' denote the fully fine-tuning paradigm and the adapter-tuning paradigm, respectively.}
	\label{tab:4}
    \scalebox{1.18}{
    	\begin{tabular}{lcllllllllll}
		\toprule
			\multirow{2}[2]{*}{\textbf{Method}} & \multirow{2}[2]{*}{\textbf{Backbone}}  & \multicolumn{2}{c}{\textbf{UCF101}} & \multicolumn{2}{c}{\textbf{HMDB51}} & \multicolumn{2}{c}{\textbf{Kinetics}} & \multicolumn{2}{c}{\textbf{SSv2-Small}}\\
      \cmidrule(lr){3-4} \cmidrule(lr){5-6} \cmidrule(lr){7-8} \cmidrule(lr){9-10}
			& & \textbf{1-shot} & \textbf{5-shot} & \textbf{1-shot} & \textbf{5-shot} & \textbf{1-shot} & \textbf{5-shot} & \textbf{1-shot} & \textbf{5-shot}\\
			\midrule
       \dag BiMACL \cite{guo2024bi} & ResNet-50 & 83.9  & 97.2  & 57.0  & 78.4  & 68.1  & 87.6 & 38.6 & 61.1\\
      \dag  SA-CT \cite{zhang2023importance} & ResNet-50 & 85.4  & 96.4  & 60.4  & 78.3  & 71.9  & 87.1  & -  & - \\
      \dag  SA-CT \cite{zhang2023importance} & IN21K-VIT-B & -  & 98.0 & - & 81.6  & - & 91.2  & -  & - \\
      \dag  CLIP-FSAR \cite{wang2024clip} & CLIP-ViT-B & 96.6  & 99.0  & 75.8  & 87.7  & \underline{89.7}  & 95.0  & 54.5  & 61.8  \\
	  \ddag ST-Adapter \cite{pan2022st} & CLIP-ViT-B & 95.9*  & 98.1* & 76.7* & 86.0*  & 88.1*  & 93.7*  & 48.3*  & 62.7*  \\
        \ddag AIM \cite{yang2023aim} & CLIP-ViT-B & 96.7*  & 98.8* & 75.9* &87.7*  & 89.2*  & 94.4*  & 50.5*  & 64.2* \\
	  \ddag $\mathrm{D}^{\mathrm{2}}$ST-Adapter \cite{pei2023d} & CLIP-ViT-B & 96.4  & \underline{99.1}  & 77.1  & \underline{88.2}  & 89.3  & \underline{95.5}  & \underline{55.0}  & \bfseries{69.3} \\
      \midrule
      \rowcolor{gray!20}
      \ddag DMGAL-Adapter (Bi-MHM) & CLIP-ViT-B & \bfseries{97.2}  &  99.0 & \bfseries{78.4} & 87.5  & \bfseries{90.1}  & 95.4  & \bfseries{55.6}  & 68.4 \\
      \rowcolor{gray!20}
	   \ddag DMGAL-Adapter (TRX) & CLIP-ViT-B & \underline{96.9}  &  \bfseries{99.2} & \underline{77.3} & \bfseries{88.8}  & 89.5  & \bfseries{95.7}  & 54.7  & \underline{69.0}\\
	\bottomrule
	\end{tabular}}
\end{table*}%

\subsection{Comparison with State-of-the-Art Methods}
\subsubsection{Results on DMGAL-FT} We compare our DMGAL-FT with existing state-of-the-art methods across different datasets under the 5-way K-shot setting, where K is typically 1, 3, or 5. We employ four commonly used matching metrics to evaluate the effectiveness of our DMGAL-FT, including OTAM \cite{cao2020few}, SCA \cite{zhang2023importance}, Bi-MHM \cite{wang2022hybrid}, and TRX \cite{perrett2021temporal}. The results demonstrate that our model outperforms a substantial number of prior works in the 1-shot, 3-shot and 5-shot setting across all datasets.

For SSv2-Small and SSv2-Full datasets that focus more on temporal context, the proposed method significantly improves performance. Table \ref{tab:2} presents a comparison of state-of-the-art methods on the SSv2-Small and SSv2-Full datasets using DMGAL-FT. For instance, DMGAL-FT outperforms the four matching metrics by a significant margin of at least 3.4\% in the 5-way 1-shot task and achieves an average improvement of 6.4\% in the 5-way 5-shot task on SSv2-Small. The representative task-specific method HyRSM \cite{wang2022hybrid} leverages class prototypes for learning task-specific features. However, it fails to account for the spatio-temporal relationships between different videos at the task level. Compared to HyRSM using Bi-MHM \cite{wang2022hybrid} as the matching metric, DMGAL-FT (Bi-MHM) improves the performance from 40.6\% to 42.8\%, achieving the best results under the 1-shot SSv2-Small setting. 

The same phenomenon is more significant for the SSv2-Full dataset, DMGAL-FT brings a large improvement across different matching metrics under the 5-way K-shot setting. The TRX \cite{perrett2021temporal} matching metric utilizes a cross-attention mechanism to match all sub-sequences of support and query videos. Although TRX achieves competitive results in the 5-way 5-shot task on SSv2-Full, its performance in the 1-shot task is unsatisfactory. By combining DMGAL to learn the spatio-temporal relationships from video-specific to task-specific, DMGAL-FT (TRX) significantly improves performance from 42.0\% to 55.5\% under the 1-shot setting and achieves a 6.5\% improvement under the 5-shot setting. The above experimental results on the SSv2-Small and SSv2-Full temporal datasets demonstrate that the proposed method can explicitly capture short-term and long-term motion information to enhance temporal relation modeling. This also indicates that learning the spatio-temporal relationships from the video-specific level to the task-specific level is crucial for understanding the temporal context in few-shot action recognition.

For spatial-related datasets such as UCF101, HMDB51, and Kinetics, which primarily focus on appearance, the proposed method consistently outperforms state-of-the-art methods, as shown in Table~\ref{tab:3}. The reason behind this is that when S-MGA and C-MGA extract motion patterns, they actually enhance the spatial information of the motion-related regions from videos to the task. We can surpass the competitive method CCLN \cite{wang2024cross} on UCF101 by 1.1\% in the 5-way 1-shot task and by 1.3\% in the 5-way 5-shot task. The proposed method shows consistent improvements across both spatial and temporal datasets using four matching metrics. For instance, DMGAL-FT (TRX) achieves the best results with 97.4\% on UCF101, 79.3\% on HMDB51, and 88.0\% on Kinetics in the 5-way 5-shot task. These results confirm that modeling spatio-temporal relationships of motion-related regions guided by motion is crucial for understanding the spatial context in few-shot action recognition. 

In the fine-tuning paradigm, DMGAL-FT shows competitive results compared with state-of-the-art methods across both spatial and temporal datasets. The experimental results demonstrate the effectiveness and importance of using S-MGA and C-MGA as additional modules to enable the model to sequentially learn spatio-temporal relationships at the video level and the task level.

\begin{figure*}[t]
	\centering
	\includegraphics[width=\linewidth]{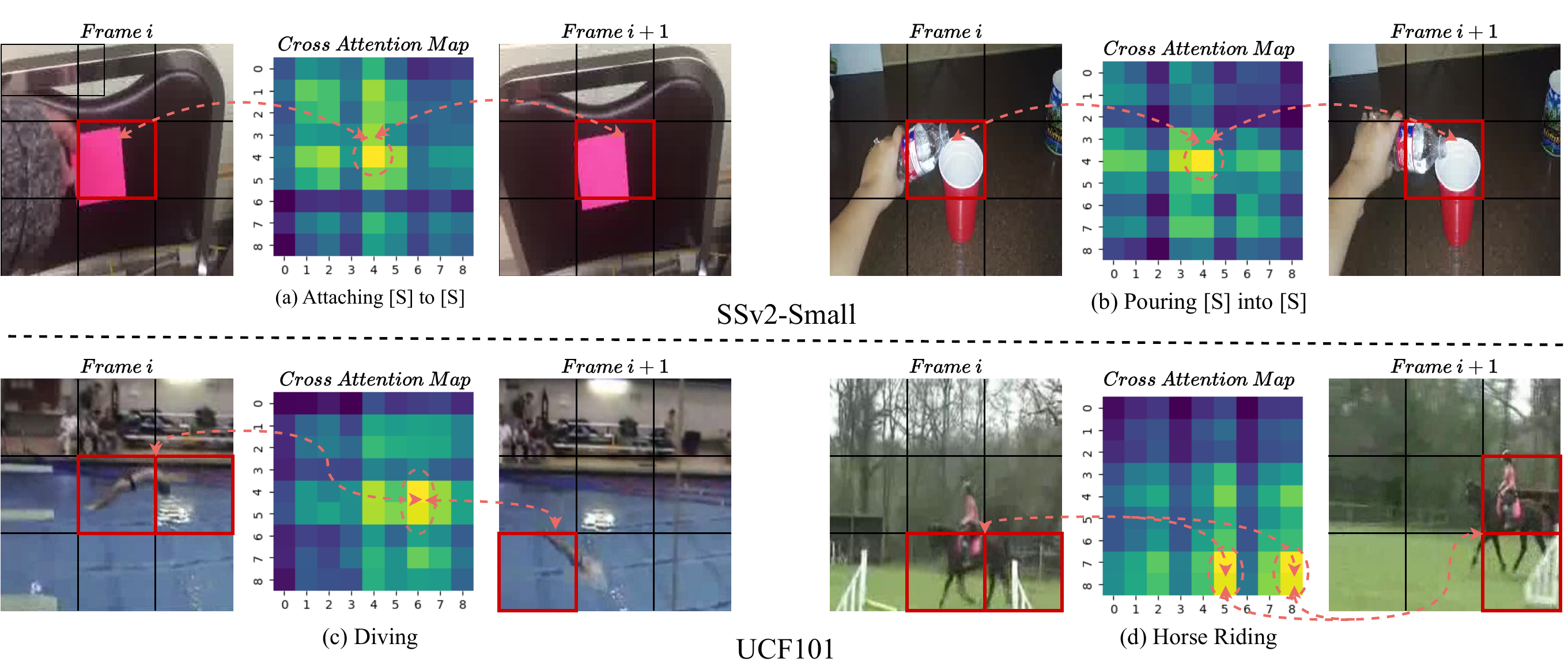}
	\caption{Visualization of the cross-association ability of S-MGA on four examples using the UCF and SSv2-small datasets. For better visualization, we downsample the total number of patches to $3 \times 3 = 9$ patches. The vertical axis represents the patches of $F_i$ ($3 \times 3 = 9$ grid flattened to 9 patches), while the horizontal axis represents the patches of $F_{i+1}$ ($3 \times 3 = 9$ grid flattened to 9 patches). Brighter colors indicate higher similarity.}
	\label{fig:4}%
\end{figure*}

\subsubsection{Results on DMGAL-Adapter} 
Simply fine-tuning the pre-trained model can cause overfitting due to the scarcity of video samples. However, recent works \cite{pan2022st, pei2023d} employ adapter-tuning to achieve parameter-efficient learning of spatio-temporal features. This approach enables a more efficient transfer of knowledge from image foundation models \cite{li2023blip, radford2021learning} to video models, leading to improved performance. Recognizing its potential for few-shot action recognition, we have designed the DMGAL-Adapter model specifically for adapter-tuning. We compare the DMGAL-Adapter with state-of-the-art methods on the UCF101, HMDB51, Kinetics, and SSv2-Small datasets to demonstrate its effectiveness. The results are shown in Table~\ref{tab:4}.

The proposed method still exhibits competitive performance across different datasets when employing the adapter-tuning paradigm. For example, compared to the ST-Adapter, the proposed method shows an overall improvement (from 0.1\% to 4.3\%) across all datasets. Although the DMGAL-Adapter shows a slight disadvantage compared to the D$^{\mathrm{2}}$ST-Adapter in the 5-shot setting on SSv2-Small, its improvement from 55.0\% to 55.6\% in the 1-shot setting remains competitive. The experimental results demonstrate the effectiveness and importance of enabling the model to sequentially learn spatio-temporal relationships at both the video and task levels within the adapter-tuning paradigm.

\begin{table}[t]
	\centering
	\caption{The influence of the proposed modules on UCF101 and SSv2-Full using DMGAL-FT (TRX). We chose TRX as the matching metric.}
	\scalebox{1.15}{
		\begin{tabular}{cccccc}
			\toprule
			\multicolumn{1}{c}{\multirow{2}[2]{*}{\textbf{S-MGA}}}  & \multirow{2}[2]{*}{\textbf{C-MGA}} & \multicolumn{2}{c}{\textbf{UCF101}} & \multicolumn{2}{c}{\textbf{SSv2-Full}} \\
      \cmidrule(lr){3-4} \cmidrule(lr){5-6}
			&      &  \textbf{1-shot} & \textbf{5-shot} & \textbf{1-shot} & \textbf{5-shot} \\
			\midrule
			\ding{56}  & \ding{56}  & 78.2 & 96.1 & 42.0  & 64.6 \\
			\ding{52}  & \ding{56}  & 85.6 & 97.2 & 52.4  & 69.2 \\
			\ding{56}  & \ding{52}  & 85.0 & 96.9 & 49.7  & 68.1 \\
            \rowcolor{gray!20}
			\ding{52}  & \ding{52}  & \bfseries{87.1} & \bfseries{97.4} & \bfseries{55.5} & \bfseries{71.1} \\
			\bottomrule
			
		\end{tabular}%
	}
	\label{tab:5}%
\end{table}%

\subsection{Ablation study}
\subsubsection{Influence of the components of DMGAL-FT} 
We decompose our proposed DMGAL-FT model and conduct a detailed evaluation of each component. The two key components of DMGAL-FT are S-MGA and C-MGA, which are used for spatio-temporal relation modeling for video-to-task learning in the fully fine-tuning paradigm. As shown in Table~\ref{tab:5}, we first compare the performance gains of S-MGA and C-MGA when applied separately. When S-MGA is applied individually, the performance of 1-shot on UCF101 and SSv2-Full can be significantly improved by 7.4\% and 10.4\%, respectively. Using C-MGA alone also results in large enhancements of 6.8\% and 7.7\%, respectively. These results demonstrate the effectiveness of S-MGA and C-MGA. Subsequently, we observe a further improvement in few-shot action recognition performance when S-MGA and C-MGA are used together. Specifically, the performance of 1-shot on UCF101 improves from 78.2\% to 87.1\%, and on SSv2-Full, it improves from 42.0\% to 55.5\% compared to the baseline TRX. This suggests that gradually learning features related to video-level and task-level motion-related regions is beneficial for understanding spatial-temporal relationships, thereby contributing to reliable video matching.

\subsubsection{Influence of the components of DMGAL-Adapter}
Different from DMGAL-FT, DMGAL-Adapter is used to conduct spatio-temporal relation modeling for video-to-task learning in the adapter-tuning paradigm. The key components of DMGAL-Adapter are Smga-Adapter and Cmga-Adapter, which are simplified variants of S-MGA and C-MGA as shown in Figure~\ref{fig:3}(b). The evaluation of each model component is presented in Table~\ref{tab:6}. The first row presents the results of using Bi-MHM as the matching metric for few-shot action recognition with the ST-Adapter, which is our baseline in the adapter-tuning paradigm. We observe that replacing the ST-Adapter with the Smga-Adapter results in performance improvements of 3.2\% and 1.8\% for 1-shot tasks on HMDB51 and SSv2-Small, respectively. Additionally, there is a slight improvement in the 5-shot setting on both datasets. Finally, introducing the Cmga-Adapter to enhance the spatio-temporal relationships from video to task, the performance of DMGAL-Adapter can be further improved in both the 1-shot and 5-shot settings. The ablation study results demonstrate the effectiveness of the Smga-Adapter and Cmga-Adapter, emphasizing the crucial role of motion-guided attention in the adapter-tuning paradigm for video-to-task learning.

\begin{table}[t]
	\centering
	\caption{The influence of the proposed modules on HMDB51 and SSv2-Small using DMGAL-Adapter (Bi-MHM). We chose Bi-MHM as the matching metric. Smga-A. and Cmga-A. denote Smga-Adapter and Cmga-Adapter, respectively.}
	\scalebox{1.14}{
		\begin{tabular}{cccccc}
			\toprule
			\multicolumn{1}{c}{\multirow{2}[2]{*}{\textbf{Smga-A.}}}  & \multirow{2}[2]{*}{\textbf{Cmga-A.}} & \multicolumn{2}{c}{\textbf{HMDB1}} & \multicolumn{2}{c}{\textbf{SSv2-Small}} \\
      \cmidrule(lr){3-4} \cmidrule(lr){5-6}
			&      &  \textbf{1-shot} & \textbf{5-shot} & \textbf{1-shot} & \textbf{5-shot} \\
			\midrule
			\ding{56}  & \ding{56}  & 74.1 & 87.3 & 53.1 & 68.0 \\
			\ding{52}  & \ding{56}  & 77.3 & 87.4 & 54.9 & 68.3 \\
            \rowcolor{gray!20}
			\ding{52}  & \ding{52}  & \bfseries{78.4} & \bfseries{87.5} & \bfseries{55.6} & \bfseries{68.4} \\
			\bottomrule
			
		\end{tabular}%
	}
	\label{tab:6}%
\end{table}%

\begin{figure*}[t]
	\centering
	\includegraphics[width=\linewidth]{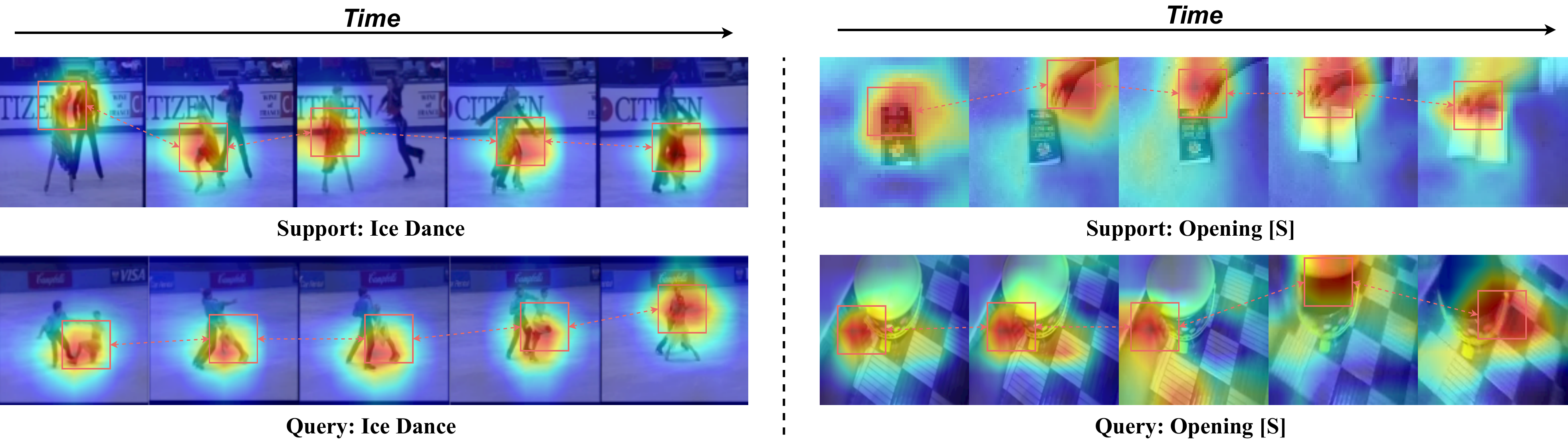}
	\caption{The attention map visualization of S-MGA identifies and correlates motion-related regions at the video level. The left side displays the attention map visualization for the UCF dataset, and the right side for the SSv2-small dataset.}
	\label{fig:5}
\end{figure*}

\begin{figure*}[t]
	\centering
	\includegraphics[width=\linewidth]{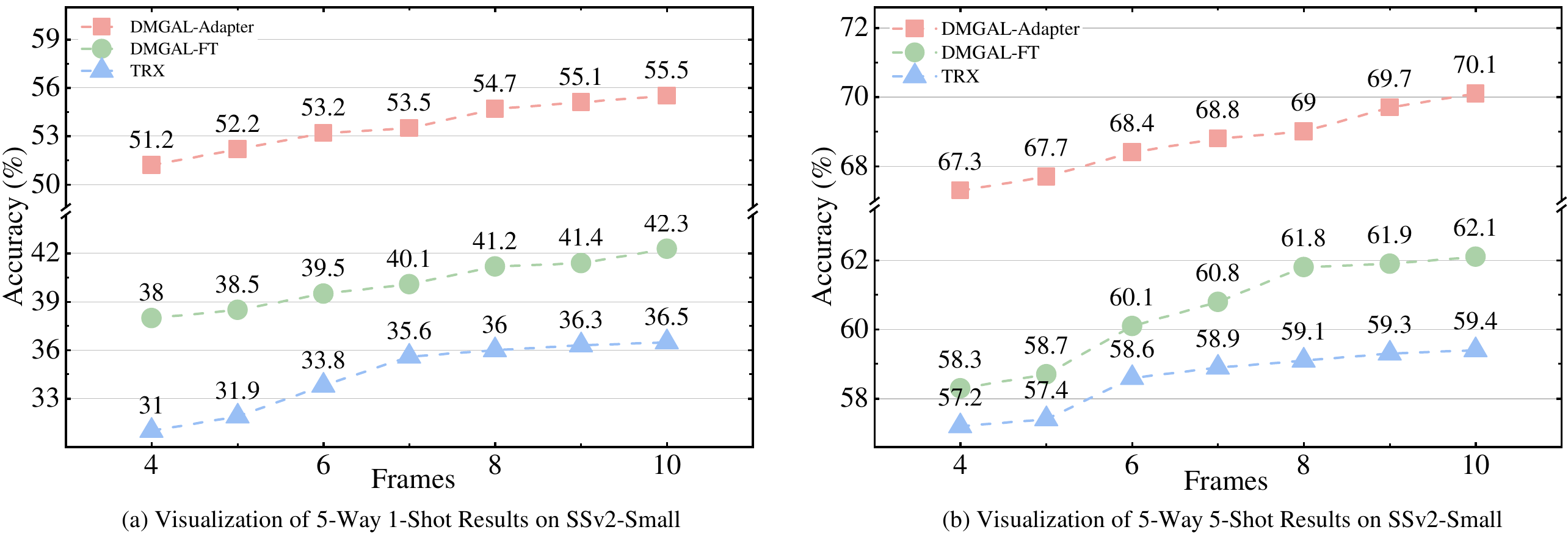}
	\caption{Ablation study on the effect of the number of input frames under the 5-way 1-shot and 5-way 5-shot settings on the SSv2-Small dataset.}
	\label{fig:6}
\end{figure*}

\subsubsection{Effect of the number of input frames}
For a fair comparison with current methods \cite{perrett2021temporal}, our proposed DMGAL-FT/DMGAL-Adapter uniformly sampled 8 frames for each video as input. To quantitatively analyze the effect of the number of input frames, we test the few-shot performance by sampling different numbers of input frames ranging from 4 to 10. From Fig.~\ref{fig:6}, we can observe that the performance of both our proposed DMGAL-FT/DMGAL-Adapter and the baseline improves rapidly as the number of input frames initially increases. However, the improvement of the baseline decelerates later due to visual information redundancy. In contrast, our proposed DMGAL-FT/DMGAL-Adapter consistently maintains stable improvement. We attribute this to the fact that an increase in the number of frames provides more delicate motion-related information, leading to more discriminative motion-related region features from videos to the task.

\begin{figure*}[t]
	\centering
	\includegraphics[width=0.85\linewidth]{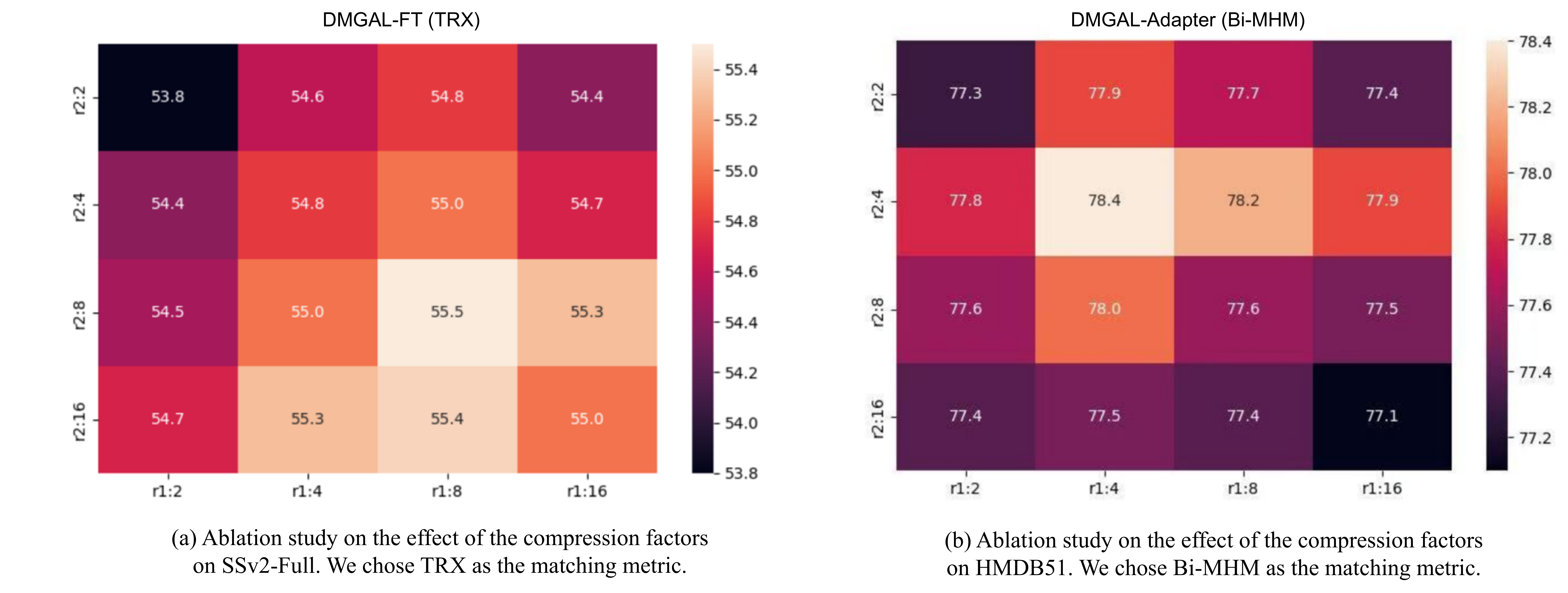}
	\caption{Ablation study on the effect of the compression factors $r_1$ and $r_2$ in the 5-way 1-shot setting. We conducted experiments on DMGAL-FT (left) and DMGAL-Adapter (right), respectively.}
	\label{fig:7}
\end{figure*}

\subsubsection{Effect of the compression factors $r_1$ and $r_2$}
Both factors $r_1$ in S-MGA and $r_2$ in C-MGA are important hyperparameters. As shown in Figure~\ref{fig:7}, we test the few-shot performance by sampling different values of $r_1$ and $r_2$ under the 5-way 1-shot SSv2-Full setting using DMGAL-FT (TRX) and the 5-way 1-shot HMDB51 setting using DMGAL-Adapter (Bi-MHM). Different values of $r_1$ and $r_2$ result in S-MGA and C-MGA having different numbers of parameters. For instance, Table~\ref{tab:7} shows the impact of different values of $r_2$ on the parameters of C-MGA. We observe that when both $r_1$ and $r_2$ are 8, DMGAL-FT (TRX) achieves the best results in the fully fine-tuning paradigm, while DMGAL-Adapter (Bi-MHM) achieves the best results with a value of 4 in the adapter-tuning paradigm. Both models have a small number of parameters, for instance, when $r_2$ is set to 8, C-MGA has only 1.79M parameters. The experimental results indicate that using the compression factors can enhance the efficiency and effectiveness of MGA in identifying and correlating motion-related regions, ultimately reducing the number of learnable parameters while improving the performance of the model.

\begin{table}[t]
	\centering
	\caption{The number of parameters in C-MGA for different values of $r_2$.}
	\begin{tabular}{ccccc}
		\toprule
		C-MGA & $r_2$:2  & $r_2$:4  & $r_2$:8  & $r_2$:16 \\
		\midrule
		Parm   & 22.06M & 6.06M & 1.79M & 0.59M \\
		\bottomrule
	\end{tabular}%
	\label{tab:7}%
\end{table}%

\begin{figure*}[t]
	\centering
	\includegraphics[width=\linewidth]{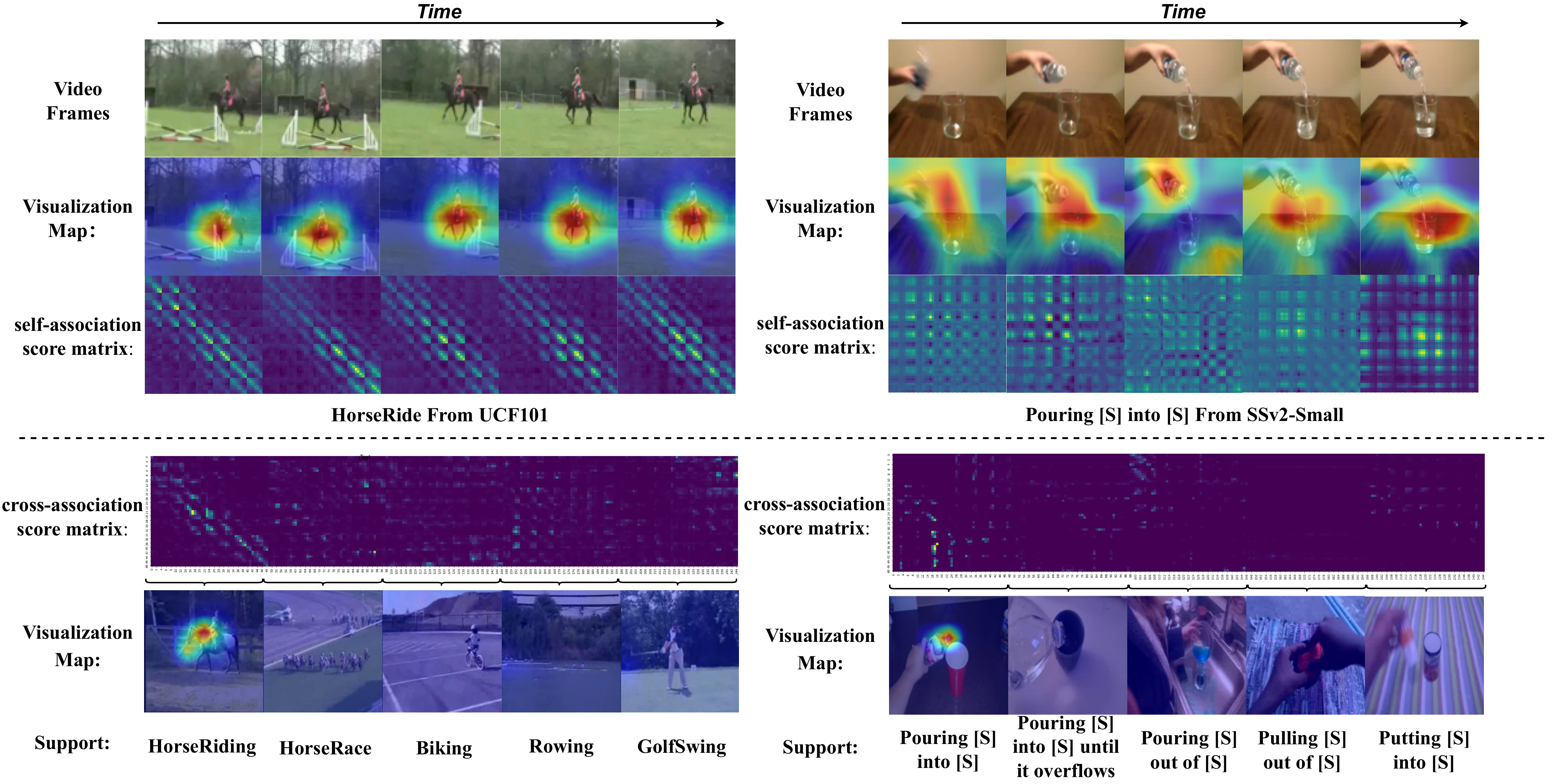}
	\caption{Attention map visualization of S-MGA and C-MGA. The brighter the colors, the higher the similarity. Results show that our MGA successfully optimizes the correlation between motion-related regions in videos and tasks.}
	\label{fig:8}
\end{figure*}

\begin{figure}[t]
	\centering
	\includegraphics[width=\linewidth]{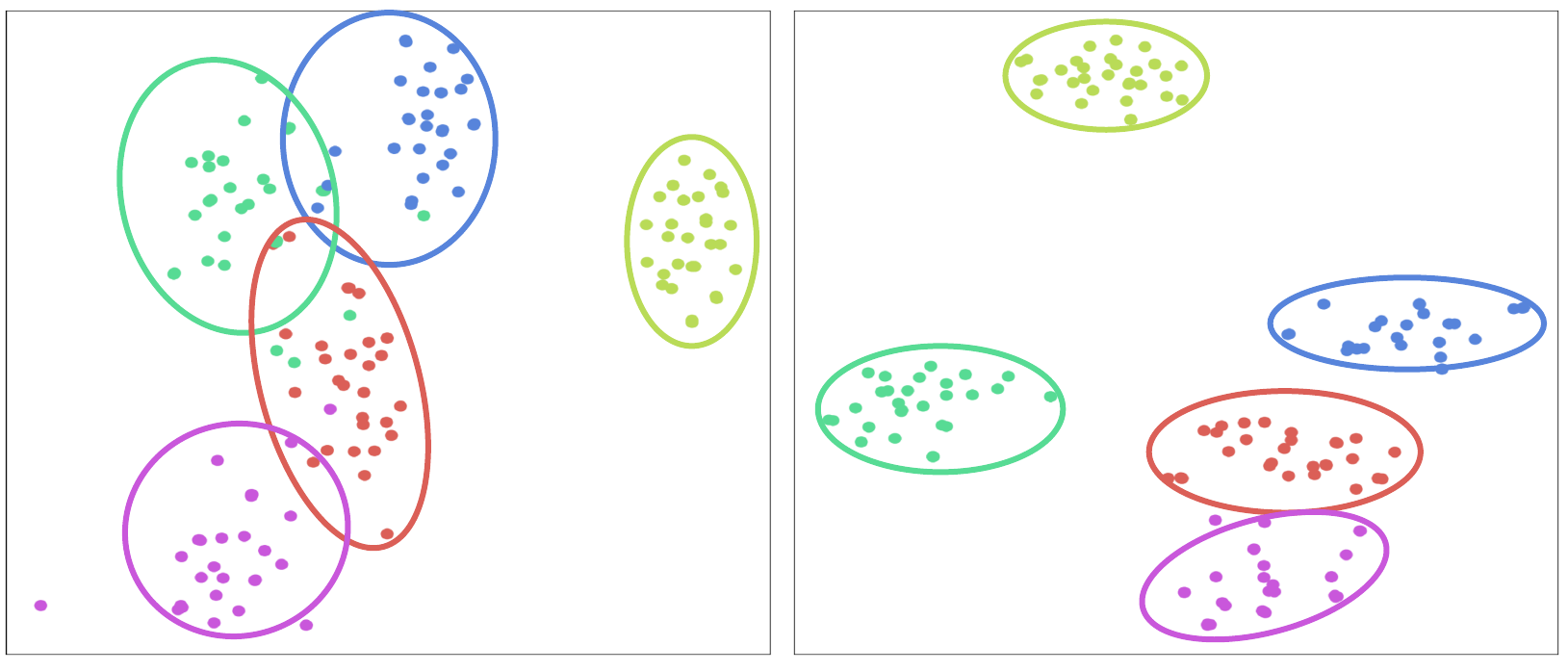}
	\caption{t-SNE feature embeddings of TRX without (left) and with (right) C-MGA, as reported using all videos in a 5-way 30-shot episode on HMDB51.}
	\label{fig:9}
\end{figure}

\subsubsection{Effect of C-MGA cross-attention computation method}
One of the main that C-MGA faces in modeling spatio-temporal relations at the task level is the difficulty of learning sparse spatio-temporal correspondences in few-shot scenarios. In our task-specific method, we address this issue through frame-wise cross motion-guided attention. Different from frame-wise computation method, frame-all computation method calculates motion-related cross-association score matrices using all frames of the entire support set in relation to the query set, resulting in a significant memory overhead. We compared these two methods applied to both the C-MGA module and the DMGAL-FT (TRX) model. Results indicated that the frame-wise method actually achieves superior results with lower memory overhead, as shown in Table~\ref{tab:8}. We attribute this to the fact that the frame-wise method calculation leads to a denser cross-association score matrix along the spatio-temporal domain, which alleviates overly sparse spatio-temporal correspondences in few-shot scenarios. Both experimental results on the C-MGA module and the DMGAL-FT model demonstrate that frame-wise cross motion-guided attention is effective and efficient.

\begin{table}[t]
	\centering
	\caption{Ablation study on the effect of the C-MGA cross-attention computation method on SSv2-Full and UCF101 in the 5-way 1-shot and 5-way 5-shot settings. We chose TRX as the matching metric.}
	\scalebox{1.08}{
    \begin{tabular}{llcccc}
		\toprule
		\multicolumn{2}{c}{\multirow{2}[2]{*}{\textbf{Method}}} & \multicolumn{2}{c}{\textbf{UCF101}} & \multicolumn{2}{c}{\textbf{SSv2-Full}} \\
        \cmidrule(lr){3-4} \cmidrule(lr){5-6}
		& & \textbf{1-shot} & \textbf{5-shot} & \textbf{1-shot} & \textbf{5-shot} \\
		\midrule
		\multicolumn{2}{l}{C-MGA (frame-all)} & 84.4  & 96.5  & 48.1  & 67.4 \\
        \rowcolor{gray!20}
		\multicolumn{2}{l}{C-MGA (frame-wise)} & 85.0  & 96.9  & 48.7  & 67.9 \\
		\multicolumn{2}{l}{DMGAL-FT (frame-all)} & 85.8  & 97.1  & 55.2  & 70.6 \\
        \rowcolor{gray!20}
		\multicolumn{2}{l}{DMGAL-FT (frame-wise)} & \textbf{87.1}  & \textbf{97.4}  & \textbf{55.5}  & \textbf{71.1} \\
		\bottomrule
	\end{tabular}}%
	\label{tab:8}%
\end{table}%

\subsection{Visualization Analysis}
\subsubsection{Visualization of the cross-association ability of S-MGA} 
In S-MGA, we employ self-attention with motion features to obtain the motion-related self-association score matrix of frame-level features $F_i$ in order to explicitly convey the degree of relevance of the motion-related regions with adjacent frames. Because the motion features contain rich motion patterns, they can naturally capture the correspondence between motion-related regions of adjacent frames implicitly. We apply cross-attention to the motion features of adjacent frames $F_i$ and $F_{i+1}$ to validate this point, as shown in Figure~\ref{fig:4}. Brighter colors indicate higher similarity.

\subsubsection{The attention map visualization of S-MGA identifies and correlates motion-related regions at the video level}
The motivation behind designing S-MGA is to capture spatio-temporal relationships within a video. To achieve this, we use self motion-guided attention module to identify and correlate motion-related regions in a video-specific manner. As shown in Figure~\ref{fig:5}, the attention map visualization of S-MGA is consistent with our motivation. S-MGA enables the model to continuously focus on and correlate motion-related regions most relevant to its category, providing a deeper understanding of the video data. For instance, S-MGA can continuously focus on the female ice dancer in Figure~\ref{fig:5} due to its cross-association ability.

\subsubsection{Attention map visualization of S-MGA and C-MGA}
We provide the attention map visualization of S-MGA and C-MGA obtained from our proposed DMGAL-FT on SSv2-Small and UCF101 in Figure~\ref{fig:8}. For the Self Motion-Guided Attention module (S-MGA), we utilize the motion-related self-association score matrix to identify and correlate motion-related regions at video level. We present not only the visualization map but also the self-association score matrix, which highlights the most motion-related regions. Furthermore, after adding the Cross Motion-Guided Attention module (C-MGA), model can identify and correlate motion-related regions at the task level. The cross-association score matrix in Figure~\ref{fig:9} represents the cross-attention matrix between the query video and the entire support set for a specific frame. We can observe that C-MGA can clearly capture the motion-related regions among similar categories. For instance, in the five categories we carefully selected, which have similar sub-actions, C-MGA accurately focuses attention on motion-related regions. The visualization results indicate that this approach maximizes the distinction between motion-related regions of different categories and minimizes the differences within the same category, ensuring a more precise task-specific mechanism.

\subsubsection{t-SNE visualization}
To further illustrate the effect of our proposed C-MGA, we demonstrate that it can increase inter-class variations while reducing intra-class variations more intuitively at the feature level. For this purpose, we visualize the feature embeddings of the query with and without C-MGA using the t-SNE \cite{van2008visualizing} method in Fig.~\ref{fig:9}. We chose TRX as our base model and performed experiments on the HMDB51 dataset. By using our C-MGA, we observe that each cluster forms a more compact structure with clearer boundaries. This further proves that C-MGA can maximize the distinction between motion-related regions of different categories and minimize the differences within the same category, thereby achieving more accurate classification.

\section{Conclusion}
In this paper, we present a novel Dual Motion-Guided Attention Learning method (called DMGAL) for few-shot action recognition. DMGAL utilizes our carefully designed Self Motion-Guided Attention module (S-MGA) and Cross Motion-Guided Attention module (C-MGA) to achieve spatio-temporal relation modeling at the video level and the task level by identifying and correlating motion-related region features. To validate the efficiency and effectiveness of our proposed DMGAL, we develop two models: DMGAL-FT using the fully fine-tuning paradigm and DMGAL-Adapter using the adapter-tuning paradigm. Both models are designed based on S-MGA and C-MGA, allowing them to gradually learn the correlation of motion-related regions from the video-specific level to the task-specific level. Extensive experiments on five few-shot action recognition benchmarks demonstrate the  effectiveness and efficiency of the proposed method.

\bibliographystyle{IEEEtran}
\bibliography{References}

\vfill
\end{document}